\documentclass{article}

\PassOptionsToPackage{numbers, compress}{natbib}
\usepackage[preprint]{neurips_2026}

\usepackage[utf8]{inputenc} % allow utf-8 input
\usepackage[T1]{fontenc}    % use 8-bit T1 fonts
\usepackage{hyperref}       % hyperlinks
\usepackage{url}            % simple URL typesetting
\usepackage{booktabs}       % professional-quality tables
\usepackage{amsfonts}       % blackboard math symbols
\usepackage{nicefrac}       % compact symbols for 1/2, etc.
\usepackage{microtype}      % microtypography
\usepackage{xcolor}         % colors
\usepackage{amsmath}
\usepackage{amssymb}
\usepackage{mathtools}
\usepackage{amsthm}
\usepackage{multirow}
\usepackage{makecell}

\usepackage[utf8]{inputenc} % allow utf-8 input
\usepackage[T1]{fontenc}    % use 8-bit T1 fonts
\usepackage{hyperref}       % hyperlinks
\usepackage{url}            % simple URL typesetting
\usepackage{booktabs}       % professional-quality tables
\usepackage{amsfonts}       % blackboard math symbols
\usepackage{nicefrac}       % compact symbols for 1/2, etc.
\usepackage{microtype}      % microtypography
\usepackage{xcolor}         % colors
\usepackage{algorithm}
\usepackage{algpseudocode}
\usepackage{multirow}
\usepackage{amsmath}
\usepackage{amssymb}
\usepackage{graphicx}
\usepackage{amsfonts}
\usepackage{amssymb}
\usepackage{bbm}
\usepackage{dsfont}
\usepackage{subcaption}
\usepackage{wrapfig}
\usepackage{placeins}  

\title{FERMI: Exploiting Relations for Membership Inference Against Tabular Diffusion Models}

\author{%
  Abtin Mahyar$^{1,2}$ \quad
  Masoumeh Shafieinejad$^{2}$ \quad
  Yuhan Liu$^{1}$ \quad
  Xi He$^{1,2}$\thanks{Corresponding author.} \\[0.5em]
  $^{1}$University of Waterloo \quad
  $^{2}$Vector Institute \\
  \texttt{\{amahyar, yuhan.liu, xi.he\}@uwaterloo.ca} \quad
  \texttt{mshafieinejad@vectorinstitute.ai}
}

\begin{document}

\maketitle

\begin{abstract}
Diffusion models are the leading approach for tabular data synthesis and are increasingly used to share sensitive records. Whether they actually protect privacy has become a pressing question. Membership inference attacks are the standard tool for this purpose, yet existing attacks assume a single-table setting and ignore the multi-relational structure of real sensitive data. A core challenge in assessing privacy risks from membership inference attacks in multi-table settings is how to leverage auxiliary information from relations associated with the target table, such as its parent tables. Particularly, we study a practical setting in which such auxiliary information is available only when training the attack model. At inference time, the attacker observes only the attribute values of the target record from the target table. We propose FERMI (FEature-mapping for Relational Membership Inference), which resolves this gap by enriching single-table features with relational membership signal. Across three tabular diffusion architectures and three real-world relational datasets, FERMI consistently improves attack performance over single-table baselines, with TPR@$0.1$FPR rising by up to 53\% over the single-table baseline in the white-box setting and 22\% in the black-box setting.

\end{abstract}
\section{Introduction}
\label{sec:intro}

Tabular data is pervasive across high-stakes domains such as healthcare \cite{giuffre2023harnessing}, finance \cite{meldrum2025new}, and social science \cite{bender2020privacy}, yet the records it contains are often highly sensitive, and privacy regulations such as GDPR and HIPAA restrict the use and direct release of individual-level data \cite{tovino2016hipaa}. To enable data sharing and analysis without exposing real records, organizations have increasingly turned to synthetic data generation. Diffusion models have emerged as the leading paradigm for tabular synthesis due to their state-of-the-art fidelity \cite{ho2020denoising, kotelnikov2023tabddpm, mendes2025synthetic}. The credibility of this approach, however, depends on whether the synthetic data truly protects privacy in practice. Membership inference attacks (MIAs) have emerged as the common empirical tool for assessing this risk, probing whether a trained generative model leaks information about its training records \cite{shokri2017membership, zhang2022membership}. Yet existing attacks against tabular generative models \cite{ward2025ensembling, van2023membership} almost exclusively assume a single-table setting, overlooking that sensitive tabular data typically resides in relational databases where an individual's information spans tens or hundreds of interconnected tables \cite{pang2024clavaddpm}, and where an adversary may exploit this broader context to mount a far more informed attack \cite{ward2026finding}. Evaluations that ignore this structure risk substantially underestimating privacy exposure, motivating the need for MIA frameworks for the multi-relational setting.

A central question in this setting is whether an adversary can leverage relational context to mount a more successful attack against tabular diffusion models. This is non-trivial: relational structure is heterogeneous across schemata, and it is not obvious how an attacker should incorporate information from linked tables. Additionally, in realistic deployments, the adversary may have accumulated rich auxiliary data with full relational structure through prior access, public releases, or breaches at related institutions. At membership inference time, however, they typically hold only a candidate row from the target table, without access to that individual's records in any related parent or child tables. For example, an adversary auditing a synthetic demographic dataset about patients may hold leaked relational records (e.g., prescriptions, lab results) for other individuals while possessing only a query row for the target patient.

We address this gap with a \textit{feature mapping framework} for membership inference against tabular diffusion models. We first show that when diffusion models are queried on merged multi-table representations, the gap in denoising loss between member and non-member records widens compared to the single-table regime, exposing a sharper and more exploitable membership signal. Using this, we then show that even when relational data is unavailable at membership inference time, an adversary who has access to auxiliary data containing relational context during the attack training phase can still benefit from this richer signal by learning a mapping that distills it into the single-table feature space. At membership inference time, the adversary operates entirely on the single-table model, applying the learned mapping to recover the stronger relational signal without requiring access to the related tables or the merged-table models for the queried row. Our main contributions are listed as follows:
\begin{enumerate}

    \item We show that relational context can yield substantially stronger membership signals in tabular diffusion models, and formalize membership inference in the multi-relational setting.

    \item We propose FERMI, a framework that distills relational membership signals into the single-table feature space, enabling stronger attacks without relational
    access at membership inference time.
    
    \item We evaluate FERMI under both white-box and black-box attack scenarios, across three diffusion architectures and three real-world relational datasets, showing consistent gains over single-table baselines.

\end{enumerate}

\section{Related Work}
\label{sec:related}

\paragraph{Tabular Data Synthesis.} Denoising diffusion probabilistic models (DDPMs) \cite{ho2020denoising} have emerged as a powerful class of generative models, achieving state-of-the-art performance across diverse data modalities. This success has motivated a growing body of work adapting diffusion-based approaches to structured tabular data. Representative methods include TabDDPM~\cite{kotelnikov2023tabddpm}, a discrete-time Gaussian diffusion framework for single-table synthesis; TabSyn \cite{zhang2023mixed}, which applies score-based diffusion in a learned VAE latent space; and TabDiff \cite{shi2024tabdiff}, a unified continuous-time formulation that jointly models numerical and categorical features through mixed-type denoising. These advances have more recently inspired efforts to extend diffusion-based synthesis to the multi-relational setting, where data is distributed across multiple interdependent tables linked via foreign-key constraints. Early work in this direction include ClavaDDPM \cite{pang2024clavaddpm}, which employs a clustering-based strategy to model parent-child relationships, and RGCLD \cite{hudovernik2024relational}, which leverages graph neural networks to encode relational structure within a latent diffusion framework.

\paragraph{Membership Inference Attacks.} MIAs were first studied against classification models \cite{shokri2017membership} and later extended to generative models \cite{hayes2017logan}, with the threat model varying by the level of access assumed \cite{ha2021comprehensive}. The shadow-model paradigm \cite{shokri2017membership} and Likelihood Ratio Attacks \cite{carlini2022membership}, which compare loss distributions across shadow models, remain standard baselines, and several approaches leverage public auxiliary data to calibrate thresholds and improve practical performance \cite{bertran2023scalable, lassila2025practical}. The shift to diffusion models has motivated tailored attacks: SecMI \cite{duan2023diffusion} exploits step-wise posterior estimation errors specific to the diffusion forward process, while other work has explored gradient-based thresholding \cite{hu2023membership} and extraction attacks that recover training examples from diffusion models \cite{carlini2023extracting}. These attacks, however, are all developed in the vision domain, targeting image diffusion models.

Despite growing interest in MIAs against tabular diffusion models, the multi-relational setting that characterizes real sensitive data has received almost no dedicated treatment. The MIDST challenge \cite{shafieinejad2026midst}, the first and most comprehensive public benchmark for tabular diffusion MIAs, included multi-table tracks with relational data, yet no submission, including the winning Tartan Federer (TF) entry \cite{wu2025winning}, made meaningful use of related-table information. The closest prior work is \cite{ward2026finding}, which represents the database as a graph and applies heterogeneous graph neural networks to aggregate a candidate's full relational subgraph, establishing that relational structure carries a valuable membership signal in the multi-table setting. FERMI builds on this insight and extends it to a strictly broader operating regime: even when the relational neighborhood of the query record is not available at membership inference time and the adversary observes only a row from the target single table, FERMI still recovers a substantial portion of the relational signal by using the knowledge gained from auxiliary data during the attack's training phase. FERMI is, to our knowledge, the first attack to formalize and resolve the training-inference asymmetry, achieving strong attack performance with no relational access at membership inference time.

\section{Background}
\label{sec:background}
\paragraph{Multi-relational Tabular Database.} A multi-relational tabular database is a collection $\mathcal{D} = \{T_1, \ldots, T_M\}$ of $M$ tables. Each table has a primary key $k_j$ used as a row identifier and excluded from generative modeling and attack features. If $T_i$ contains a foreign key referencing the primary key of $T_j$, we write $T_i \to T_j$. This induces a directed acyclic graph (DAG) over tables, where edges point from child tables to parent tables. We denote by $\text{pa}(j)$ the set of parent tables of $T_j$.

\paragraph{Tabular Diffusion Models.} We consider diffusion models trained to capture the distribution of a single table $T_j$ \cite{kotelnikov2023tabddpm}. Let $x_0^{(j)}$ denote a row from $T_j$. The forward process gradually corrupts the data by injecting Gaussian noise over $T$ discrete timesteps:

\begin{equation}
    q\left(x_t^{(j)} \mid x_0^{(j)}\right) = \mathcal{N}\left(x_t^{(j)};\, \sqrt{\bar{\alpha}_t}\, x_0^{(j)},\, (1-\bar{\alpha}_t)I\right), 
\end{equation}

where $\alpha_t = 1 - \beta_t \in (0,1)$ is determined by a small variance schedule $\{\beta_t\}_{t=1}^{T}$, and $\bar{\alpha}_t = \prod_{s=1}^{t}\alpha_s$ denotes the cumulative noise factor. The reverse process $p_{\theta_j}(x_{t-1}^{(j)} \mid x_t^{(j)})$ is parameterized by a neural network trained to predict the injected noise $\epsilon \sim \mathcal{N}(0,I)$ via the denoising objective:

\begin{equation}
\mathcal{L}_{\text{diff}}(\theta_j) = \mathbb{E}_{x_0^{(j)},\, t,\, \epsilon}\left[\left\|\epsilon - \epsilon_{\theta_j}\left(x_t^{(j)}, t\right)\right\|_2^2\right],
\end{equation}

with $x_t^{(j)} = \sqrt{\bar{\alpha}_t}\,x_0^{(j)} + \sqrt{1 - \bar{\alpha}_t}\,\epsilon$, where $t$ is drawn uniformly from $\{1,\dots,T\}$.

\paragraph{Membership Inference Attack.}  MIAs target a trained model $\mathcal{M}$ and ask, for a given record $\mathbf{x}$, whether $\mathbf{x}$ was used during training. We denote the true membership label of a candidate record $\mathbf{x}$ by $y \in \{0,1\}$, where $y = 1$ if $\mathbf{x}$ was used to train $\mathcal{M}$ and $y = 0$ otherwise. The adversary's goal is to produce a scalar membership score $r(\mathbf{x}) \in [0,1]$ that approximates $y$, with values close to $1$ indicating likely membership. A binary membership prediction is obtained by thresholding $r(\mathbf{x})$ at some $\tau$; since the choice of $\tau$ trades off true and false positive rates, attacks are evaluated across the full range of thresholds via the ROC curve, with particular emphasis on the true positive rate at low false positive rates~\cite{carlini2022membership}. The adversary's capabilities are characterized by the level of access assumed over $\mathcal{M}$: in the \emph{black-box} setting only the model's outputs (typically its synthetic samples) are observable, whereas in the \emph{white-box} setting the adversary additionally has access to the model parameters $\theta$, enabling the construction of substantially more discriminative signals. While $\mathcal{D}_{\text{train}}$ is unavailable, the adversary is typically assumed to hold an auxiliary dataset $\mathcal{D}_{\text{aux}}$ drawn from the same distribution. Prior work commonly assumes $\mathcal{D}_{\text{aux}} \cap \mathcal{D}_{\text{train}} = \emptyset$, which represents the worst case for the adversary; any overlap between the two would strengthen the attack \cite{shokri2017membership}.

The dominant strategy for constructing $r(\mathbf{x})$ under this threat model is the shadow-model paradigm \cite{shokri2017membership}, in which a collection of shadow models $\{\mathcal{M}^{(s)}\}_{s=1}^{S}$ is trained on disjoint subsets of $\mathcal{D}_{\text{aux}}$, yielding records with known membership labels. A classifier $f_{\text{att}}$ is then trained on features extracted from shadow models and used to score candidate records against the target model at membership inference time.

Following the TF attack \cite{wu2025winning}, membership signals from a diffusion model can be summarized as a loss fingerprint over multiple timesteps and noise realizations. For a row $\mathbf{x}_0$ from table $T_j$, a set of probed timesteps $\mathcal{T} = \{t_1, \ldots, t_{n_t}\}$, and noise vectors $\mathcal{E} = \{\boldsymbol{\epsilon}_1, \ldots, \boldsymbol{\epsilon}_{n_\epsilon}\}$, the feature extractor is \begin{equation}
\Phi(\mathbf{x}_0) = \left[\ell_{\theta_j}(\mathbf{x}_0, t, \boldsymbol{\epsilon})\right]_{t \in \mathcal{T},\, \boldsymbol{\epsilon} \in \mathcal{E}} \in \mathbb{R}^d, \label{eq:feature-extractor} \end{equation} 

with $d = n_t \times n_\epsilon$ and per-sample denoising loss $\ell_{\theta_j}(\mathbf{x}_0, t, \boldsymbol{\epsilon}) = \|\boldsymbol{\epsilon} - \epsilon_{\theta_j}(\mathbf{x}_t, t)\|_2^2$. Empirically, training members exhibit systematically lower loss values than non-members.

\section{Problem Formulation and Threat Model}
\label{sec:problem}

We study row-level membership inference against tabular diffusion models in multi-relational settings.\footnote{A summary of all notation used in the paper is provided in Appendix~\ref{app:notation}.} Consider a multi-relational database instance $\mathcal{D}_{\text{train}} = \{T_1, \dots, T_M\}$, and let $T_j \in \mathcal{D}_{\text{train}}$ denote the target training table on which the diffusion model $p_{\theta_j}$ is trained. Given a candidate row $\mathbf{x}$ drawn from the same underlying distribution as the rows of $T_j$, the adversary's goal is to determine whether $\mathbf{x}$ was used to train $p_{\theta_j}$, by producing a scalar membership score $r(\mathbf{x}) \in [0,1]$ that approximates true membership label $y$.

Throughout Sections~\ref{sec:problem}--\ref{sec:experiments} we consider a white-box adversary with access to (i) the target model parameters $\theta_j$, (ii) an auxiliary dataset $\mathcal{D}_{\text{aux}}$ drawn from the same distribution as $\mathcal{D}_{\text{train}}$, with no overlapping records, and (iii) the candidate row $\mathbf{x}$ from the target table $T_j$ but \emph{not} its relational neighborhood, i.e., the adversary does not have access to the parent or child records linked to $\mathbf{x}$ in $\mathcal{D}$. These assumptions are standard for white-box MIA~\cite{zhang2022membership}, with the exception of (iii), which deliberately restricts the adversary's view of the target record to a single row from $T_j$. This asymmetry between training-phase and membership inference time access is what FERMI is designed to address. Section~\ref{sec:blackbox} relaxes the white-box assumption and extends the attack to a black-box adversary that observes only the released synthetic data sampled from the diffusion model.

To study the impact of relational auxiliary information on attack performance, we consider two progressively richer settings, distinguished by the relational context available in $\mathcal{D}_{\text{aux}}$ during the attack's training phase. In both, the adversary is restricted to single-table access at membership inference time. 
\begin{itemize} 
\item \textbf{Parent-Context Side Information (PCSI).} $\mathcal{D}_{\text{aux}}$ contains the target table $T_j$ and a parent table $T_p \in \mathrm{pa}(j)$, following the same schema and distribution as $\mathcal{D}_{\text{train}}$. 
\item \textbf{Full-Relational Side Information (FRSI).} $\mathcal{D}_{\text{aux}}$ contains the full relational structure of the database, including $T_j$ and all linked relations (parents, children, and connected tables). 
\end{itemize}

\section{FERMI}
\label{sec:method}

\begin{figure*}[!tbp]
    \centering
    \includegraphics[
        width=\textwidth,
        trim=3.5cm 0.75cm 3.5cm 0.5cm, 
        clip
    ]{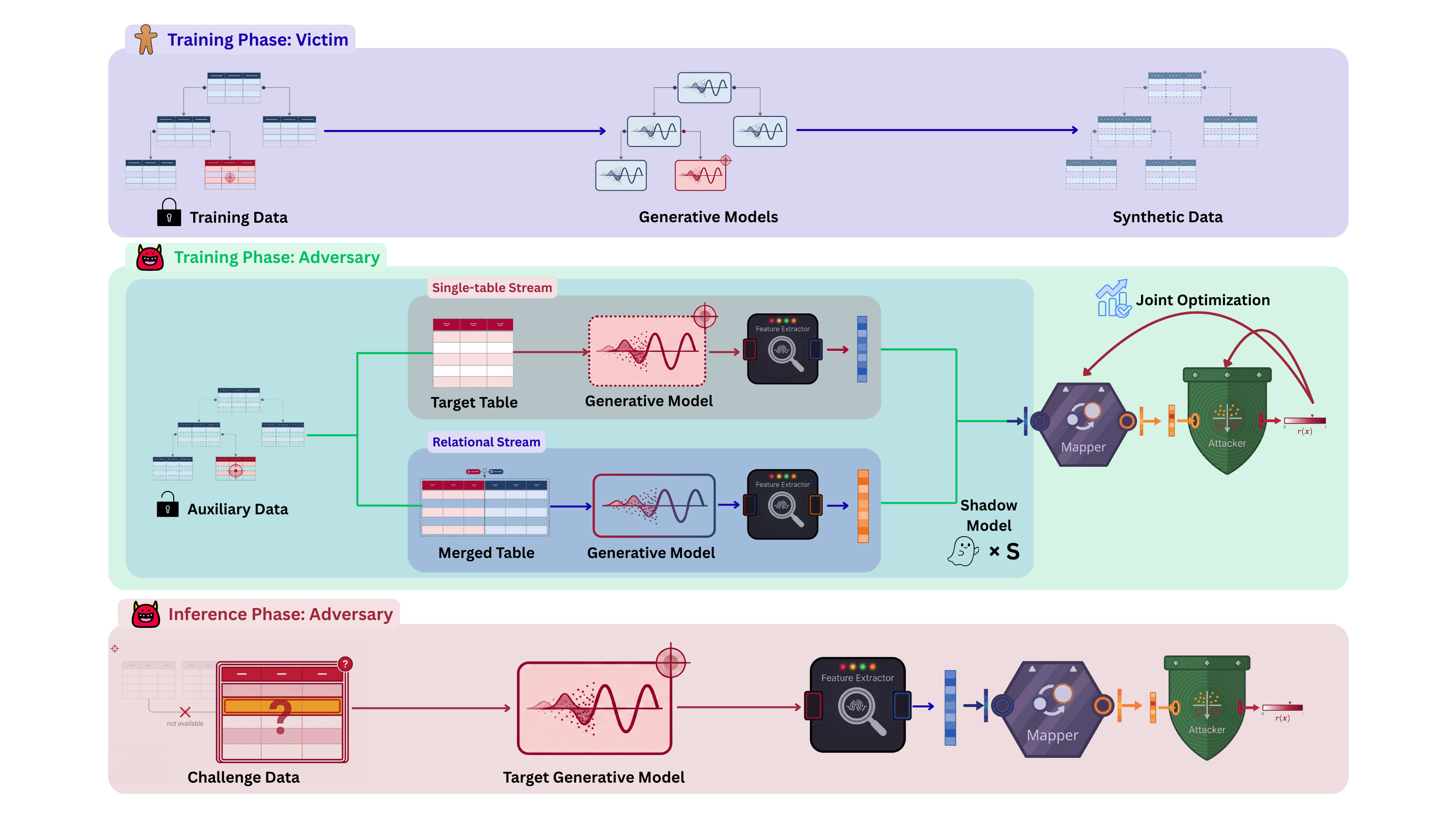}
\caption{Overview of FERMI. The victim trains per-table diffusion
models (top). The adversary extracts paired single-table and
relational features from shadow models to train the mapper and attacker (middle). At
inference, only single-table features are required (bottom).}
\label{fig:method}\end{figure*}

At a high level, FERMI infers membership by learning a mapping between two types of features: those extracted from the target table alone, and those that would have been extracted had the target table been augmented with relational context. The latter is constructed by joining the target table with related tables, such as its parent table. Overall, we follow the standard shadow-model paradigm commonly used in prior membership inference work~\cite{shokri2017membership, jayaraman2019evaluating}. As illustrated in Figure~\ref{fig:method}, on the victim side, a separate diffusion model $p_{\theta_j}$ is trained on each table $T_j \in \mathcal{D}_{\text{train}}$, and either the trained models or the synthetic data they produce are released. On the adversary side, the adversary trains $S$ shadow diffusion models $\{M^{(s)}\}_{s=1}^{S}$ on disjoint subsets of the auxiliary dataset $\mathcal{D}_{\text{aux}}$, with known membership labels for each shadow record.

FERMI proceeds in three stages. (1)~\textbf{Paired Feature Extraction.} For each record $x \in T_j$, the adversary extracts two feature vectors: $\Phi_{\text{single}}(x)$ from a shadow model trained on the target table alone, and $\Phi_{\text{multi}}(\tilde{x})$ from one trained on the same record augmented with relational context $\tilde{x}$. (2)~\textbf{Mapping.} The adversary trains a mapper $f_{\text{map}}$ to project single-table features toward their relational counterparts such that $f_{\text{map}}(\Phi_{\text{single}}(x)) \approx \Phi_{\text{multi}}(\tilde{x})$. (3)~\textbf{Membership Prediction.} The adversary trains an attack classifier $f_{\text{att}}$ on the mapped features to predict membership. At membership inference time, the adversary observes only the candidate row $x$, extracts $\Phi_{\text{single}}(x)$, applies $f_{\text{map}}$ to recover a synthetic relational view $\hat{x}_m = f_{\text{map}}(\Phi_{\text{single}}(x))$, and predicts membership via $f_{\text{att}}(\hat{x}_m)$.

\paragraph{Paired Feature Extraction.} For each shadow record $x \in T_j$ with membership label $y \in \{0,1\}$, the two streams are defined as follows:
\begin{itemize}
    \item \textbf{Single-table Stream.} Takes $x \in T_j$ as input and returns $\Phi_{\text{single}}(x)$, computed by applying the loss-fingerprint extractor $\Phi$ (Eq.~\ref{eq:feature-extractor}) to a diffusion model trained on $T_j$.
    \item \textbf{Relational Stream.} Takes the augmented record $\tilde{x} = x \bowtie_{T_p \in \mathcal{R}} x_p$ as input, where $\mathcal{R} = \{T_p\}$ for $T_p \in \mathrm{pa}(j)$ under PCSI and $\mathcal{R}$ is the relational neighborhood of $T_j$ under FRSI, and returns $\Phi_{\text{multi}}(\tilde{x})$, by applying $\Phi$ to a model trained on the merged representation. 
\end{itemize}

Intuitively, features extracted from the relational stream yield \emph{stronger membership signals} than those from single-table inputs, as they capture richer relational dependencies. Although performing such joins on large-scale relational databases can be expensive, within the shadow training phase the auxiliary partitions are small enough to make the join operation tractable. In addition, FERMI is attack-agnostic: any feature extractor derived from the diffusion process (e.g., score-based features or intermediate activations) is admissible, and the choice may differ between streams. We use the loss-fingerprint extractor $\Phi$ from Section~\ref{sec:background} for both streams.

\paragraph{Mapping.} Mapping $\Phi_{\mathrm{single}}$ to $\Phi_{\mathrm{multi}}$ can be naturally cast as a domain adaptation problem with source distribution $\Phi_{\mathrm{single}}$ and target distribution $\Phi_{\mathrm{multi}}$~\cite{singhal2023domain}. The mapper is trained on tuples $\big(\Phi_{\mathrm{single}}(x), \Phi_{\mathrm{multi}}(\tilde{x}), y\big)$ across all shadow models, with an objective combining point-wise reconstruction and distributional alignment:

\begin{equation}
\label{eq:loss_mapper}
\mathcal{L}_{\mathrm{map}} = \mathbb{E}\left[\left\| f_{\mathrm{map}}(\Phi_{\mathrm{single}}(x)) - \Phi_{\mathrm{multi}}(\tilde{x})\right\|_2^2\right] + \lambda_{\mathrm{coral}} \cdot \frac{1}{4d^2} \left\| C_{\mathrm{map}} - C_{\mathrm{multi}} \right\|_F^2,
\end{equation}

where $C_{\mathrm{map}}$ and $C_{\mathrm{multi}}$ are covariance matrices of mapped and true multi-table features. The first term enforces point-wise reconstruction; the second, implemented via CORAL~\cite{sun2017correlation}, matches the covariance structure of the two feature spaces. Together, these terms encourage the mapper to recover not only the values but also the relational geometry of the multi-table feature space.

\paragraph{Membership Prediction.} The attack classifier $f_{\mathrm{att}}$ is trained on mapped features with binary cross-entropy loss $\mathcal{L}_{\mathrm{cls}}$. After independent pretraining of $f_{\mathrm{map}}$ and $f_{\mathrm{att}}$, we jointly fine-tune both components:

\begin{equation}
    \mathcal{L}_{\mathrm{total}} = 
\lambda_{\mathrm{map}} \mathcal{L}_{\mathrm{map}} 
+ \lambda_{\mathrm{cls}} \mathcal{L}_{\mathrm{cls}}
\end{equation}

Joint optimization allows gradients from the classification objective to refine the mapper toward features most informative for membership inference. 

\paragraph{Theoretical guarantees.} An attack classifier trained directly on $\Phi_{\text{multi}}(\tilde{x})$ serves as a natural upper bound on FERMI, since FERMI's mapped features are trained to approximate this merged-table representation. We empirically verify that $\Phi_{\text{multi}}$ carries stronger membership signal than $\Phi_{\text{single}}$ in all reported settings. Formal statements and proofs are deferred to Appendix~\ref{sec:theory}.

\section{Experiments}
\label{sec:experiments}
\subsection{Setup}
\paragraph{Datasets.} We evaluate our framework on three real-world multi-relational tabular datasets, California \cite{series2020international}, Instacart \cite{instacart2017}, and Berka \cite{berka2000guide}, each characterized by diverse schemata and complex inter-table dependencies. All datasets are preprocessed to remove identifier columns and to ensure consistent relational integrity across tables. Additional details about the datasets are provided in Appendix \ref{app:datasets}.

\paragraph{Generative Models.} We evaluate our proposed membership inference attack framework against three representative tabular data generative models: 1) TabDDPM \cite{kotelnikov2023tabddpm}, a diffusion model that operates directly in the data space using discrete-time Gaussian diffusion. Notably, our results also extend to ClavaDDPM \cite{pang2024clavaddpm}, which builds upon their enhanced version of TabDDPM for multi-relational data by incorporating classifier-guided synthesis based on parent-child clustering. This modification affects only the sampling procedure and does not alter the underlying diffusion model training. Hence, the behavior of the learned denoising model, and consequently the membership inference signals, remains unchanged, and our analysis applies directly; 2) TabSyn \cite{zhang2023mixed}, a two-stage generative model that first encodes tabular records into a latent space via a variational autoencoder (VAE) and subsequently performs score-based latent diffusion on the resulting embeddings; and 3) TabDiff \cite{shi2024tabdiff}, a unified continuous-time diffusion model that applies Gaussian denoising for numerical columns and absorbing-state (masked) diffusion for categorical columns, operating directly on preprocessed tabular features, providing a comprehensive testbed for our attack framework.

While the overall attack framework, namely, extracting per-record diffusion loss features, training an attacker, and learning a cross-table feature mapper, remains consistent across all three generative models, the feature extraction procedure is adapted to each model’s diffusion formulation. We provide a detailed description of these adaptations for each model in Appendix \ref{app:model_adaptations}, and full hyperparameter and training details in Appendix \ref{app:hyper}.

\paragraph{Evaluation Metrics.} We evaluate attack performance using standard membership inference metrics. Following common practice in the literature \cite{carlini2022membership}, we focus on metrics that better capture the effectiveness of an attack under realistic operating conditions, where low false positive rates are critical. Accordingly, our primary metric is the true positive rate at low false positive rates (TPR@FPR), reported at two operating points $\mathrm{FPR}=0.1$ and $\mathrm{FPR}=0.01$. In addition, we report the area under the ROC curve (AUC-ROC) as a threshold-independent measure of performance.

\paragraph{End-to-end Evaluation.} For each generative model, we train 10 independent instances on disjoint partitions of the same relational auxiliary dataset to serve as shadow models. For these models, training membership labels are available to the adversary and are used to train the attack classifier. At membership inference time, the attack is evaluated on 5 target models, each trained on disjoint and independent data partitions that are not used during the attack training phase. We report the mean performance across these 5 target models, along with standard deviations.

For each model instance, we construct a balanced challenge set of 200 records, consisting of 100 members randomly sampled from the model’s training data and 100 non-members sampled from held-out test data. Additionally, non-member pools are deduplicated using record-level key columns to ensure no overlap with any model’s training data or challenge records across the full set of models.

We evaluate our framework under three progressively informative adversarial settings, corresponding to increasing levels of auxiliary knowledge:

\begin{itemize}
    \item \textbf{Direct (Single-Table).} The adversary has access only to a single-table generative model trained on the target table. Loss-based features are extracted from these models and directly used to train the attack classifier. This setting represents a minimal-knowledge baseline.
    
    \item \textbf{Direct (Merged-Table).} The adversary has access to a generative model trained on a merged (denormalized) table that augments the child table with relevant parent-table attributes. Features extracted from these models incorporate richer relational context and are used to train the attacker. This setting evaluates whether additional structural information strengthens the membership signal.
    
    \item \textbf{FERMI.} The adversary is restricted to single-table models at training and inference time, but aims to leverage the stronger signals present in merged-table representations from the auxiliary sets. To this end, a learned feature mapper projects single-table loss features into the merged-table feature space, followed by an attack classifier trained on the mapped representations. At membership inference time, only the single-table model and the learned mapper are required, with no need for access to a merged-table model. This setting assesses whether cross-table feature transfer can enhance attack performance without direct access to multi-relational models at test time.
\end{itemize}

All three generators achieve strong utility and fidelity, so differences in MIA susceptibility reflect architectural properties rather than generative quality; full metrics are reported in Appendix~\ref{app:utility}. All experiments were conducted on a NVIDIA L40 GPU and 16 CPU cores.

\setlength{\intextsep}{0pt}
\begin{wraptable}{r}{0.49\textwidth}
\centering
\footnotesize
\setlength{\tabcolsep}{3pt}
\caption{Direct single-table membership inference performance (baseline). Values are mean $\pm$ std.}
\label{tab:single}
\begin{tabular}{ll|ccc}
\toprule
Dataset & Model & AUC & TPR@0.1 & TPR@0.01 \\
\midrule
\multirow{3}{*}{Berka}
& TabDDPM & .753\,{\scriptsize$\pm$.031} & .388\,{\scriptsize$\pm$.043} & .266\,{\scriptsize$\pm$.023} \\
& TabDiff & .659\,{\scriptsize$\pm$.037} & .258\,{\scriptsize$\pm$.012} & .180\,{\scriptsize$\pm$.043} \\
& TabSyn  & .503\,{\scriptsize$\pm$.045} & .136\,{\scriptsize$\pm$.026} & .024\,{\scriptsize$\pm$.010} \\
\midrule
\multirow{3}{*}{Instacart}
& TabDDPM & .949\,{\scriptsize$\pm$.010} & .806\,{\scriptsize$\pm$.040} & .636\,{\scriptsize$\pm$.062} \\
& TabDiff & .619\,{\scriptsize$\pm$.052} & .178\,{\scriptsize$\pm$.027} & .034\,{\scriptsize$\pm$.015} \\
& TabSyn  & .494\,{\scriptsize$\pm$.045} & .108\,{\scriptsize$\pm$.012} & .030\,{\scriptsize$\pm$.019} \\
\midrule
\multirow{3}{*}{California}
& TabDDPM & .896\,{\scriptsize$\pm$.022} & .640\,{\scriptsize$\pm$.046} & .456\,{\scriptsize$\pm$.092} \\
& TabDiff & .806\,{\scriptsize$\pm$.070} & .512\,{\scriptsize$\pm$.107} & .208\,{\scriptsize$\pm$.113} \\
& TabSyn  & .518\,{\scriptsize$\pm$.055} & .112\,{\scriptsize$\pm$.055} & .010\,{\scriptsize$\pm$.009} \\
\bottomrule
\end{tabular}
\end{wraptable}

\subsection{Results}

\paragraph{Single-table Baseline.} Table \ref{tab:single} reports attack performance under a minimal-knowledge baseline, wherein the adversary operates solely on single-table diffusion models without access to relational context. TabDDPM exhibits the highest susceptibility to membership inference across all three datasets, indicating that its data-space Gaussian diffusion process retains significant amounts of per-record training information. TabDiff's vulnerability, in contrast, varies across datasets, strong on California but near random on Instacart. This pattern suggests that dataset-specific characteristics, such as the prevalence of categorical versus numerical columns, interact with TabDiff's mixed-type diffusion formulation to differentially influence membership leakage.

\begin{table*}[!bp]
\centering
\footnotesize
\setlength{\tabcolsep}{6pt}
\caption{Membership inference under PCSI. Values are mean $\pm$ std.}
\label{tab:pcsi}
\begin{tabular}{ll|ccc|ccc}
\toprule
\multirow{2}{*}{Dataset} & \multirow{2}{*}{Model}
& \multicolumn{3}{c|}{Merged}
& \multicolumn{3}{c}{FERMI} \\
\cmidrule(lr){3-5} \cmidrule(lr){6-8}
& & AUC & TPR@0.1 & TPR@0.01 & AUC & TPR@0.1 & TPR@0.01 \\
\midrule
\multirow{3}{*}{Berka}
& TabDDPM & .937\,{\scriptsize$\pm$.009} & .772\,{\scriptsize$\pm$.050} & .502\,{\scriptsize$\pm$.026} & .811\,{\scriptsize$\pm$.031} & .504\,{\scriptsize$\pm$.059} & .350\,{\scriptsize$\pm$.065} \\
& TabDiff & .810\,{\scriptsize$\pm$.024} & .458\,{\scriptsize$\pm$.049} & .348\,{\scriptsize$\pm$.026} & .713\,{\scriptsize$\pm$.022} & .378\,{\scriptsize$\pm$.031} & .238\,{\scriptsize$\pm$.025} \\
& TabSyn  & .512\,{\scriptsize$\pm$.023} & .132\,{\scriptsize$\pm$.019} & .030\,{\scriptsize$\pm$.030} & .517\,{\scriptsize$\pm$.019} & .134\,{\scriptsize$\pm$.017} & .022\,{\scriptsize$\pm$.017} \\
\midrule
\multirow{3}{*}{Instacart}
& TabDDPM & .991\,{\scriptsize$\pm$.003} & .996\,{\scriptsize$\pm$.008} & .846\,{\scriptsize$\pm$.071} & .967\,{\scriptsize$\pm$.010} & .892\,{\scriptsize$\pm$.040} & .658\,{\scriptsize$\pm$.111} \\
& TabDiff & .652\,{\scriptsize$\pm$.080} & .242\,{\scriptsize$\pm$.075} & .096\,{\scriptsize$\pm$.052} & .629\,{\scriptsize$\pm$.027} & .232\,{\scriptsize$\pm$.032} & .072\,{\scriptsize$\pm$.038} \\
& TabSyn  & .564\,{\scriptsize$\pm$.032} & .166\,{\scriptsize$\pm$.060} & .052\,{\scriptsize$\pm$.065} & .520\,{\scriptsize$\pm$.019} & .134\,{\scriptsize$\pm$.020} & .024\,{\scriptsize$\pm$.010} \\
\midrule
\multirow{3}{*}{California}
& TabDDPM & .992\,{\scriptsize$\pm$.005} & .992\,{\scriptsize$\pm$.008} & .908\,{\scriptsize$\pm$.086} & .915\,{\scriptsize$\pm$.011} & .722\,{\scriptsize$\pm$.046} & .496\,{\scriptsize$\pm$.108} \\
& TabDiff & .967\,{\scriptsize$\pm$.029} & .874\,{\scriptsize$\pm$.119} & .714\,{\scriptsize$\pm$.187} & .935\,{\scriptsize$\pm$.015} & .752\,{\scriptsize$\pm$.021} & .550\,{\scriptsize$\pm$.082} \\
& TabSyn  & .574\,{\scriptsize$\pm$.047} & .144\,{\scriptsize$\pm$.060} & .036\,{\scriptsize$\pm$.035} & .529\,{\scriptsize$\pm$.039} & .144\,{\scriptsize$\pm$.039} & .018\,{\scriptsize$\pm$.018} \\
\bottomrule
\end{tabular}
\end{table*}

\begin{table*}[!tbp]
\centering
\footnotesize
\setlength{\tabcolsep}{6pt}
\caption{Membership inference under FRSI. Values are mean $\pm$ std.}
\label{tab:frsi}
\begin{tabular}{ll|ccc|ccc}
\toprule
\multirow{2}{*}{Dataset} & \multirow{2}{*}{Model}
& \multicolumn{3}{c|}{Merged}
& \multicolumn{3}{c}{FERMI} \\
\cmidrule(lr){3-5} \cmidrule(lr){6-8}
& & AUC & TPR@0.1 & TPR@0.01 & AUC & TPR@0.1 & TPR@0.01 \\
\midrule
\multirow{3}{*}{Berka}
& TabDDPM & .951\,{\scriptsize$\pm$.005} & .876\,{\scriptsize$\pm$.037} & .416\,{\scriptsize$\pm$.070} & .835\,{\scriptsize$\pm$.015} & .572\,{\scriptsize$\pm$.034} & .390\,{\scriptsize$\pm$.020} \\
& TabDiff & .814\,{\scriptsize$\pm$.014} & .432\,{\scriptsize$\pm$.039} & .076\,{\scriptsize$\pm$.033} & .722\,{\scriptsize$\pm$.017} & .394\,{\scriptsize$\pm$.033} & .272\,{\scriptsize$\pm$.059} \\
& TabSyn  & .511\,{\scriptsize$\pm$.027} & .154\,{\scriptsize$\pm$.016} & .020\,{\scriptsize$\pm$.006} & .519\,{\scriptsize$\pm$.024} & .138\,{\scriptsize$\pm$.030} & .022\,{\scriptsize$\pm$.013} \\
\midrule
\multirow{3}{*}{Instacart}
& TabDDPM & .997\,{\scriptsize$\pm$.004} & .994\,{\scriptsize$\pm$.008} & .994\,{\scriptsize$\pm$.008} & .963\,{\scriptsize$\pm$.010} & .910\,{\scriptsize$\pm$.030} & .676\,{\scriptsize$\pm$.086} \\
& TabDiff & .843\,{\scriptsize$\pm$.039} & .530\,{\scriptsize$\pm$.104} & .240\,{\scriptsize$\pm$.113} & .668\,{\scriptsize$\pm$.034} & .258\,{\scriptsize$\pm$.068} & .056\,{\scriptsize$\pm$.016} \\
& TabSyn  & .573\,{\scriptsize$\pm$.017} & .184\,{\scriptsize$\pm$.023} & .056\,{\scriptsize$\pm$.024} & .530\,{\scriptsize$\pm$.013} & .146\,{\scriptsize$\pm$.014} & .022\,{\scriptsize$\pm$.010} \\
\bottomrule
\end{tabular}
\end{table*}

TabSyn, in contrast, proves remarkably resistant to all single-table attack variants, achieving near-random AUC values across the three datasets. TPR@0.1 and TPR@0.01 values remain close to a random classifier. This could stem from TabSyn's two-stage architecture: the VAE encoder introduces a representational bottleneck that regularizes per-record memorization in the latent space, effectively absorbing membership signals before they can propagate into the diffusion component's loss landscape. Taken together, these baseline results could establish that diffusion model vulnerability is architecture-dependent, that data-space models (TabDDPM and TabDiff) are considerably more susceptible than latent-space models (TabSyn) against the applied attack.

\paragraph{Merged Attacks.} Tables \ref{tab:pcsi} and \ref{tab:frsi} report attack performance under the PCSI and FRSI settings respectively. Since California has only two tables, FRSI coincides with PCSI and is reported only under PCSI. In both cases, we consider two attack variants: the Merged setting, in which the attacker has access to a generative model trained on the denormalized table at both training and inference time, and FERMI, which restricts inference to the single-table model while exploiting relational signals during the training phase via the learned feature mapper.

The Merged attack results under PCSI demonstrate that incorporating parent-table attributes substantially amplifies membership leakage for TabDDPM and TabDiff. This confirms our central hypothesis that relational structure encodes richer conditioning signals that the denoising model internalizes during training, leaving stronger membership traces. Under FRSI, the Merged attack reaches near-ceiling performance for TabDDPM on Instacart. This remarkable result, in which virtually every true member can be identified at a 1\% false positive rate, underscores the severe privacy risk posed by diffusion models trained on relational data when an adversary has access to a fully denormalized training view. TabDiff also benefits from broader relational context under FRSI, indicating that the inclusion of related table information provides meaningful signal beyond parent context alone. TabSyn continues to resist all relational attack variants. Even with the full relational neighborhood available, AUC values remain close to random guessing levels, confirming that the latent-space bottleneck provides structural privacy protection against the applied attack that is robust to relational augmentation. This finding has important implications for architectural choices in privacy-sensitive tabular synthesis: the latent encoding appears to be the primary determinant of membership inference resistance.

\paragraph{FERMI.} Under PCSI, FERMI yields consistent and substantial improvements for TabDDPM and TabDiff across all datasets. Under FRSI, the improvements are generally larger. Comparing FERMI against the Merged upper bound provides a direct measure of how much relational signal the mapper successfully transfers. These results validate the core design premise of the framework: that diffusion-based loss features carry sufficient information for a neural mapper to distill multi-table membership signals into the single-table feature space. 

For all datasets and both side-information settings, FERMI improves TabSyn's AUC by at most +0.036 and TPR@0.01 by at most +0.008 over the single-table baseline. The residual attack performance remains statistically near chance. Unlike TabDDPM and TabDiff, where the Merged attack substantially outperforms the single-table baseline, TabSyn's Merged attack performance remains near random across all datasets and both side-information settings. Since the mapper's role is to transfer the stronger relational signal into the single-table feature space, it has nothing useful to distill when the relational upper bound itself is uninformative. This points to an important scope condition for the framework: FERMI is effective when the Merged attack meaningfully outperforms the Direct Single baseline, in other words, when the enriched relational features carry genuinely stronger membership information. This marks a limit of the current FERMI design: when the membership signal from the extracted features are near-random regardless of relational context, as is the case for TabSyn, the mapper cannot manufacture signal that does not exist in the source space. Extending the framework to such regimes is left for future investigation.

\subsection{Extension to the Black-Box Setting}
\label{sec:blackbox}

The white-box threat model assumed throughout Sections~\ref{sec:problem}--\ref{sec:experiments} grants the adversary direct access to the parameters of the target diffusion model. We now relax this assumption and ask whether the relational signal exploited by FERMI persists when the adversary observes only the synthetic data released by the target model. We evaluate this extension on TabDDPM trained on California.

\paragraph{Threat model.}
The adversary has no access to the target model's architecture, parameters, or training procedure, and observes only its released synthetic tables. Auxiliary access remains as in the white-box setting: the adversary holds a disjoint $\mathcal{D}_{\text{aux}}$ with the same relational structure as $\mathcal{D}_{\text{train}}$.

\paragraph{Attack pipeline.}
Following the TF attack \cite{wu2025winning}, to recover loss-based membership signals without parameter access, the adversary trains \emph{surrogate} diffusion models on the released synthetic data. Once trained, surrogates serve as stand-ins for the target model: per-record loss fingerprints $\Phi(\cdot)$ are extracted from them exactly as  in the white-box case, and the rest of the FERMI pipeline (single-table stream, relational stream, mapper $f_{\text{map}}$, attacker $f_{\text{att}}$, joint fine-tuning) is unchanged. The intuition is that records well-represented in the synthetic output, typically those that were members of the original training set, will be reconstructed with lower loss by a surrogate trained on that synthetic data, preserving the member--non-member separation that drives the white-box attack.

\paragraph{Results.}

\setlength{\intextsep}{0pt}
\begin{wraptable}{r}{0.55\textwidth}
\centering
\caption{Black-box MIA performance on TabDDPM trained on California. Values are mean $\pm$ std.}
\label{tab:blackbox}
\small
\setlength{\tabcolsep}{4pt}
\begin{tabular}{lccc}
\toprule
Attack & AUC & TPR\,@\,$0.1$ & TPR\,@\,$0.01$ \\
\midrule
Single-table & $.687 \pm .021$ & $.348 \pm .038$ & $.208 \pm .062$ \\
FERMI        & $.725 \pm .038$ & $.424 \pm .062$ & $.250 \pm .025$ \\
Merged       & $.755 \pm .015$ & $.496 \pm .038$ & $.374 \pm .029$ \\
\bottomrule
\end{tabular}
\end{wraptable}

Table~\ref{tab:blackbox} reports attack performance under all three variants. The qualitative ordering established in the white-box setting persists: the merged-table attack provides the strongest signal, the single-table attack the weakest, and FERMI closes a meaningful portion of the gap using only single-table access at inference time. FERMI improves AUC by $+0.038$, TPR\,@\,$0.1$ by $+0.076$ and TPR\,@\,$0.01$ by $+0.042$ over the single-table baseline, recovering roughly $55\%$ of the AUC gap, $51\%$ of the TPR\,@\,$0.1$ gap, and $25\%$ of the TPR\,@\,$0.01$ gap to the merged-table upper bound. As expected, absolute attack performance is lower than in the white-box regime, surrogate-based loss signals are noisier than direct parameter access, but the relational gap that motivates FERMI remains exploitable. This indicates that the privacy risk identified in our white-box analysis is not an artifact of strong adversary assumptions: an adversary with only the released synthetic data and auxiliary relational context can still mount a stronger attack by transferring relational signal into the single-table feature space.

\subsection{Ablation Study}
\label{sec:ablation}

We conduct two ablations to validate the design of FERMI, with full details deferred to Appendix~\ref{app:ablation}. Visualizing the per-timestep denoising loss for members and non-members shows that the mapper shifts the single-table feature distribution toward the relational regime, opening a wider member and non-member gap that concentrates at small timesteps. We further verify that this gain reflects genuine relational supervision rather than added capacity.

\section{Conclusion}
\label{sec:conclusion}
We introduced FERMI, a feature-mapping framework for membership inference against tabular diffusion models in the multi-relational setting, which enables an adversary to recover the stronger membership signal available under relational context while querying the target model with only a candidate row from a single table at membership inference time. Across three diffusion architectures and three real-world relational datasets, and under white-box and black-box adversaries, we show that relational auxiliary context at training time yields substantially stronger attacks than single-table access alone, and that FERMI closes a meaningful portion of this gap without relational access at inference time. Our experiments further reveal that, under the applied attacks, diffusion models that operate directly on the original data are substantially more vulnerable than those using a latent encoding, and this gap persists even under the richest relational augmentation. We leave latent-space–specific attacks to future work. These findings suggest that evaluating tabular diffusion models in single-table settings meaningfully underestimates true privacy risk, and that architectural decisions made upstream of privacy mechanisms deserve closer scrutiny in privacy-sensitive deployments.

\bibliographystyle{plainnat}
\bibliography{neurips_2026}

%%%%%%%%%%%%%%%%%%%%%%%%%%%%%%%%%%%%%%%%%%%%%%%%%%%%%%%%%%%%

\appendix

\section{Datasets}
\label{app:datasets}

Here we describe the three real-world multi-relational datasets used in our
evaluation. Summary statistics are reported in Table~\ref{tab:dataset-stats}.
For each dataset, we identify a single \emph{target table} on which membership
inference is performed; this table is selected as the most data-rich and
privacy-sensitive table in the schema, and corresponds to the table whose
generative model the adversary attacks.

\begin{table}[H]
\centering
\caption{Statistics of the three multi-relational datasets used in our evaluation. Depth refers to the longest directed path in the foreign-key DAG.}
\label{tab:dataset-stats}
\begin{tabular}{lccccc}
\toprule
Dataset & \#Tab. & \#FK & Depth & \#Attr. & \#Rows (target) \\
\midrule
California   & 2 & 1 & 2 & 15 & 1{,}690{,}642 \\
Instacart 05 & 6 & 6 & 3 & 12 & 1{,}616{,}315 \\
Berka        & 8 & 8 & 4 & 41 & 1{,}056{,}320 \\
\bottomrule
\end{tabular}
\end{table}

\textbf{California.} The California dataset is a real-world census database
\cite{series2020international} containing household-level information. It consists of two tables organized as a basic parent–child relationship, where the parent table
describes household-level attributes and the child table records individual-level
attributes for each member of the household. The target table for membership
inference is the \emph{individual} table, which contains the per-person records
that constitute the most sensitive layer of the schema.

\textbf{Instacart 05.} The Instacart dataset is constructed by uniformly
downsampling 5\% of the Kaggle Instacart Market Basket Analysis competition
dataset \cite{instacart2017}, a real-world record of grocery transactions. The
original schema contains 6 tables, with the order–product associations
distributed across two intermediate linking tables (one for prior orders and one
for the training split) that map each order to its constituent products. To
simplify the relational structure without discarding any information, we merge
these two linking tables into the orders table, so that each row in the
resulting orders table directly carries a foreign-key reference to the products
table. This preprocessing step reduces the effective schema to 4 tables while
preserving every original foreign-key relationship. The target table for membership inference is the \emph{order} table, which is both the largest and the most relationally connected table in the schema.

\textbf{Berka.} The Berka dataset is a real-world financial dataset
\cite{berka2000guide} that records bank account activity across a population of clients. It consists of 8 tables with a maximum schema depth of 4, making it the most
relationally complex dataset in our evaluation. The target table for membership inference is the \emph{transaction} table, which records individual financial transactions and is both the largest table and the most privacy-sensitive component of the schema.

\section{Model-Specific Adaptations and Implementation Details}
\label{app:model_adaptations}

\paragraph{TabDDPM \cite{kotelnikov2023tabddpm}} employs discrete-time Gaussian diffusion with integer timesteps. We probe the model at seven timesteps $t \in \{5, 10, 20, 30, 40, 50, 100\}$. For each record and timestep, we draw $N = 500$ independent Gaussian noise samples, forward-diffuse the record using the standard diffusion process, and compute the noise-prediction loss between the model’s denoising output and the injected noise. This results in a feature vector of dimensionality $7 \times 500 = 3{,}500$ per record. ClavaDDPM \cite{pang2024clavaddpm} additionally requires loading parent–child clustering checkpoints, which assign each child record to a cluster conditioned on its parent; this cluster assignment serves as the conditioning variable for the denoising function.

\paragraph{TabDiff \cite{shi2024tabdiff}} adopts a continuous-time diffusion process with per-column learned noise schedules. We probe the model at seven time points $t \in \{0.01, 0.05, 0.1, 0.2, 0.4, 0.6, 0.9\}$, spanning the full noise range $[0,1]$. For numerical columns, we compute the EDM-weighted denoising loss, where the weighting depends on the noise scale induced by the learned schedule. For categorical columns, we additionally compute an ELBO-weighted log-likelihood under the absorbing-state (masked) diffusion parameterization, yielding a complementary categorical loss component. When both numerical and categorical branches are used, each record yields a feature vector of dimensionality $7 \times 500 \times 2 = 7{,}000$.

\paragraph{TabSyn \cite{zhang2023mixed}} performs diffusion in a learned latent space rather than directly on raw tabular data. Each record is first encoded using a pre-trained VAE encoder into a flattened latent representation (excluding the CLS token) and normalized using the mean of the training latent distribution. We then probe the latent diffusion model at seven EDM noise scales $\sigma \in \{0.01, 0.05, 0.1, 0.2, 0.5, 1.0, 2.0\}$. For each noise level and record, we inject $N = 500$ Gaussian noise samples scaled by $\sigma$, and compute the corresponding EDM-weighted denoising loss in latent space. This yields a feature vector of dimensionality $7 \times 500 = 3{,}500$ per record. Notably, in this setting, the attack signal reflects the model’s ability to denoise latent representations rather than raw tabular values, introducing an additional representational bottleneck through which membership information must propagate.

\section{Theoretical Analysis}
\label{sec:theory}

We establish formal guarantees on the behavior of the Feature Mapping framework relative to the Direct Single-Table and Direct Merged-Table attacks. We first prove a non-degradation result that bounds the framework's performance from below, characterize the Merged attack as a natural upper bound, identify the necessary condition under which targeting the merged feature space can be useful at all, and finally clarify the role of relational supervision in driving the empirical gains observed in Section~\ref{sec:experiments}.

\paragraph{Theorem 1 (Non-Degradation).}
\label{thm:nondeg}
Let $\mathcal{H}_\text{single}$ denote the class of all measurable single-table attack functions $g: \mathbb{R}^{d_\text{single}} \to [0,1]$, and let $\mathcal{H}_\text{map}$ denote the class of all feature-mapping attack functions of the form $f_\text{att} \circ f_\text{map}$, where $f_\text{map}: \mathbb{R}^{d_\text{single}} \to \mathbb{R}^{d_\text{multi}}$ and $f_\text{att}: \mathbb{R}^{d_\text{multi}} \to [0,1]$ are measurable. Let $\mathcal{A}(\mathcal{H})$ denote the optimal attack performance achievable by any function in $\mathcal{H}$. Then
\begin{equation}
    \mathcal{A}(\mathcal{H}_\text{map}) \geq \mathcal{A}(\mathcal{H}_\text{single}).
\end{equation}

\paragraph{Proof.}
It suffices to show $\mathcal{H}_\text{single} \subseteq \mathcal{H}_\text{map}$. Let $g \in \mathcal{H}_\text{single}$ be any single-table attack. Define the embedding
\begin{equation}
    f_\text{map}^\dagger(z) = \begin{pmatrix} z \\ \mathbf{0}_{d_\text{multi} - d_\text{single}} \end{pmatrix} \in \mathbb{R}^{d_\text{multi}}, \quad z \in \mathbb{R}^{d_\text{single}},
\end{equation}
and the recovery classifier
\begin{equation}
    f_\text{att}^\dagger(z') = g(z'_{1:d_\text{single}}), \quad z' \in \mathbb{R}^{d_\text{multi}},
\end{equation}
where $z'_{1:d_\text{single}}$ denotes the first $d_\text{single}$ components of $z'$. By construction, $f_\text{att}^\dagger \circ f_\text{map}^\dagger = g$, so $g \in \mathcal{H}_\text{map}$. Therefore $\mathcal{H}_\text{single} \subseteq \mathcal{H}_\text{map}$, and
\begin{equation}
    \mathcal{A}(\mathcal{H}_\text{map}) = \sup_{h \in \mathcal{H}_\text{map}} \mathcal{A}(h) \geq \sup_{g \in \mathcal{H}_\text{single}} \mathcal{A}(g) = \mathcal{A}(\mathcal{H}_\text{single}). \tag*{$\square$}
\end{equation}

\paragraph{Remark 1 (Practical Scope of Theorem 1).}
\label{rem:scope}
Theorem~\ref{thm:nondeg} is a statement about optimal attacks over unrestricted measurable function classes, and establishes that the feature-mapping pipeline can in principle subsume any single-table attack. In practice, however, $f_\text{map}$ is trained to target $\Phi_\text{multi}(\tilde{x})$ rather than to preserve $\Phi_\text{single}(x)$, so the realized mapped features may diverge from the single-table baseline under finite shadow data. Specifically, the parent attributes introduced in $\tilde{x} = x \bowtie x_p$ induce a different denoising geometry in the merged model, which can entangle the membership signal for $x$ with relational correlations that are uninformative or even adversarial with respect to the child record's membership. In such cases, the merged feature space may carry a weaker membership signal for $m(x)$ than $\Phi_\text{single}(x)$ does in isolation, and training the mapper toward $\Phi_\text{multi}(\tilde{x})$ could yield worse finite-sample performance than the single-table baseline. The framework is therefore expected to help when the merged feature space carries at least as strong a membership signal as the single-table space, a condition that is directly verifiable by comparing the Merged attack against the Direct Single baseline before deploying the framework, and that we confirm empirically for all settings where the framework is applied in Section~\ref{sec:experiments}.

\paragraph{Remark 2 (Merged Attack as Upper Bound).}
\label{rem:upper}
Within a fixed attack-classifier family $\mathcal{F}$, the Direct Merged-Table attack constitutes a natural upper bound on the Feature Mapping framework's performance. The framework produces scores $f_\text{att}(f_\text{map}(\Phi_\text{single}(x)))$, which depend on $\Phi_\text{multi}(\tilde{x})$ only through the approximation $f_\text{map}(\Phi_\text{single}(x)) \approx \Phi_\text{multi}(\tilde{x})$. As mapping fidelity improves, i.e., $f_\text{map}(\Phi_\text{single}(x)) \to \Phi_\text{multi}(\tilde{x})$, the framework's performance with classifier $f_\text{att} \in \mathcal{F}$ approaches that of the Merged attack with the same classifier family. Since the mapper is trained to minimize the reconstruction objective $\mathcal{L}_\text{map}$ (Eq.~\ref{eq:loss_mapper}), this convergence is the explicit target of training. Together with Theorem~\ref{thm:nondeg}, this places the framework's performance within the interval $[\mathcal{A}(\mathcal{H}_\text{single}), \mathcal{A}(\mathcal{H}_\text{merged})]$, with position within that interval governed by the mapper's reconstruction quality.

\paragraph{Proposition 1 (Necessary Condition for Useful Relational Supervision).}
\label{prop:necessary}
If $\mathcal{A}(\mathcal{H}_\text{merged}) > \mathcal{A}(\mathcal{H}_\text{single})$, then $m(x)$ is not conditionally independent of $\Phi_\text{multi}(\tilde{x})$ given $\Phi_\text{single}(x)$; equivalently,
\begin{equation}
    I\bigl(m(x);\, \Phi_\text{multi}(\tilde{x}) \,\big|\, \Phi_\text{single}(x)\bigr) > 0,
\end{equation}
where $I(\cdot;\cdot \mid \cdot)$ denotes conditional mutual information.

\paragraph{Proof.} Suppose for contradiction that $m(x) \perp \Phi_\text{multi}(\tilde{x}) \mid \Phi_\text{single}(x)$. Then for any measurable $h: \mathbb{R}^{d_\text{multi}} \to [0,1]$, the function
\begin{equation}
    \bar{h}(z) := \mathbb{E}\bigl[h(\Phi_\text{multi}(\tilde{x})) \,\big|\, \Phi_\text{single}(x) = z\bigr]
\end{equation}
is a measurable function of $\Phi_\text{single}(x)$ that yields the same posterior over $m(x)$ as $h$. Hence every classifier on $\Phi_\text{multi}$ is matched, in posterior and therefore in attack performance, by some classifier on $\Phi_\text{single}$, implying $\mathcal{A}(\mathcal{H}_\text{merged}) \leq \mathcal{A}(\mathcal{H}_\text{single})$, contradicting the hypothesis. \hfill $\square$

\paragraph{Remark 3 (Relational Supervision as Inductive Bias).}
\label{rem:inductive}
At the level of optimal attacks over unrestricted function classes, the data processing inequality implies $\mathcal{A}(\mathcal{H}_\text{map}) = \mathcal{A}(\mathcal{H}_\text{single})$, since $f_\text{map}(\Phi_\text{single}(x))$ is a deterministic function of $\Phi_\text{single}(x)$ and cannot carry additional information about $m(x)$. The empirical improvements reported in Section~\ref{sec:experiments} therefore arise not from the mapper recovering information absent from $\Phi_\text{single}$, but from the relational supervision signal acting as an inductive bias on the feature representation: training $f_\text{map}$ against $\Phi_\text{multi}(\tilde{x})$ shapes the mapped features toward geometries that downstream classifiers of bounded capacity can exploit more effectively than they exploit $\Phi_\text{single}(x)$ directly. This interpretation is consistent with the identity-mapping ablation in Appendix~\ref{sec:ablation}, where removing the relational target collapses the framework to the single-table baseline despite identical model capacity and training budget, indicating that the gains are attributable to the supervision signal rather than to additional parameters or optimization steps.

\section{Hyperparameters and Training Details}
\label{app:hyper}

\subsection{Feature Extraction}
Across all three generative models, we use $N = 500$ noise samples per record at each probing point (i.e., timestep, continuous time value, or noise scale). For the 10 shadow models, we sample 1,000 member records and 1,000 non-member records per model to construct the attacker’s training set. For the 5 target models, we sample 200 members and 200 non-members per model. Member records are drawn from each model’s training split, while non-members are sampled from a disjoint set, with additional key-based deduplication to prevent overlap.

\subsection{Attacker Network}
The attacker is implemented as a multi-layer perceptron (MLP) with batch normalization, ReLU activations, and dropout. The architecture consists of 4 hidden layers with a hidden dimension of 128 and a dropout rate of 0.2. All attacker models are trained using the Adam optimizer with binary cross-entropy loss and full-batch gradient updates per epoch. The learning rate is set to $5 \times 10^{-3}$.

\subsection{Feature Mapper}
In FERMI, we employ a neural network with layer normalization, ReLU activations, and residual connections to project single-table features into the merged-table feature space. The mapper is trained using mean squared error (MSE) loss with the Adam optimizer and a scheduler. The mapper uses a hidden dimension of 128 with 3 residual blocks, trained for 100 epochs with a learning rate of $5 \times 10^{-4}$.

The feature-mapping pipeline further incorporates domain adaptation through a CORAL loss during mapper training. Specifically, we use a CORAL weight of $\lambda_{\text{coral}} = 0.01$. This addition mitigates the distributional shift between single-table and multi-table feature spaces.

\subsection{End-to-End Fine-Tuning}
Following independent pre-training of the mapper and attacker, we jointly fine-tune both components using a combined objective that balances feature reconstruction and membership classification. Specifically, the loss consists of an MSE term that aligns the mapped features with their merged-table counterparts, and a binary cross-entropy (BCE) term for membership prediction. A reconstruction weight of $\lambda_{\text{map}} = 0.05$ is used consistently across all models. End-to-end training is performed using the Adam optimizer with a scheduler. Training proceeds for 500 epochs. The learning rate is set to $2 \times 10^{-3}$.

\section{Utility and Fidelity of the Generative Models}
\label{app:utility}

To ensure that differences in MIA susceptibility across architectures
reflect architectural properties rather than disparities in generative
quality, we evaluate all three generators on five standard utility and
fidelity metrics. For each (model, dataset) pair we use $15$
independent synthesized tables and report the mean $\pm$ standard deviation
across runs.

\textbf{Shape} measures column-wise marginal fidelity, computed as the
Kolmogorov--Smirnov statistic for numerical columns and the Total
Variation Distance (TVD) for categorical columns, averaged across columns
(lower is better). \textbf{Trend} measures pair-wise correlation
fidelity, defined as the absolute Pearson correlation difference for
numerical pairs and the TVD between joint contingency tables for
categorical pairs, averaged across all pairs (lower is better).
\textbf{$\alpha$-Precision} and \textbf{$\beta$-Recall}
\citep{liu2023goggle, alaa2022faithful} measure sample-level fidelity and coverage,
respectively, using a $k$-nearest-neighbor support estimator with
$k=5$ on subsamples of $5{,}000$ records (higher is better, max $1.0$).
\textbf{C2ST} (classifier two sample test) reports the cross-validated accuracy of a logistic regression classifier trained to distinguish real from synthetic samples
(closer to $0.5$ is better) \cite{zhang2023mixed}.

Table~\ref{tab:utility} reports the results. TabSyn and TabDDPM achieve
consistently strong scores across all metrics and datasets, with
$\alpha$-Precision and $\beta$-Recall above $0.86$ and C2ST scores
within $\pm 0.03$ of the ideal $0.5$. TabDiff matches this performance
on Berka and California but exhibits a notable drop on Instacart, where
$\alpha$-Precision and $\beta$-Recall fall to roughly $0.50$ and the
C2ST score rises to $0.74$. Importantly, this degradation does not
translate into stronger MIA performance: TabDiff remains substantially
less vulnerable than TabDDPM on Instacart in
Tables~\ref{tab:pcsi}--\ref{tab:frsi}. This reinforces that MIA
susceptibility in our experiments is governed by architectural properties
of the diffusion process rather than by raw generative quality. 

\begin{table*}[!tbp]
\centering
\caption{Utility and fidelity metrics for the generative models
across the datasets. Values are mean $\pm$ std over $15$
independently generated synthetic tables. Lower is better for Shape and
Trend; higher is better for $\alpha$-Precision and $\beta$-Recall; closer
to $0.5$ is better for C2ST.}
\label{tab:utility}
\small
\setlength{\tabcolsep}{4pt}
\begin{tabular}{llccccc}
\toprule
Model & Dataset & Shape $\downarrow$ & Trend $\downarrow$ & $\alpha$-Prec.\ $\uparrow$ & $\beta$-Recall $\uparrow$ & C2ST ($\to 0.5$) \\
\midrule
\multirow{3}{*}{TabSyn}
  & Berka      & $.018 \pm .005$ & $.022 \pm .004$ & $.963 \pm .003$ & $.939 \pm .005$ & $.514 \pm .006$ \\
  & California & $.029 \pm .018$ & $.039 \pm .023$ & $.880 \pm .008$ & $.882 \pm .009$ & $.521 \pm .021$ \\
  & Instacart  & $.021 \pm .001$ & $.016 \pm .002$ & $.970 \pm .003$ & $.930 \pm .003$ & $.513 \pm .007$ \\
\midrule
\multirow{3}{*}{TabDiff}
  & Berka      & $.016 \pm .002$ & $.022 \pm .002$ & $.946 \pm .003$ & $.951 \pm .003$ & $.518 \pm .004$ \\
  & California & $.011 \pm .002$ & $.020 \pm .002$ & $.868 \pm .006$ & $.893 \pm .006$ & $.498 \pm .007$ \\
  & Instacart  & $.027 \pm .001$ & $.024 \pm .002$ & $.496 \pm .035$ & $.495 \pm .051$ & $.741 \pm .006$ \\
\midrule
\multirow{3}{*}{TabDDPM}
  & Berka      & $.017 \pm .005$ & $.022 \pm .004$ & $.963 \pm .004$ & $.937 \pm .006$ & $.513 \pm .007$ \\
  & California & $.043 \pm .021$ & $.058 \pm .025$ & $.883 \pm .008$ & $.862 \pm .008$ & $.530 \pm .020$ \\
  & Instacart  & $.024 \pm .002$ & $.018 \pm .003$ & $.970 \pm .003$ & $.912 \pm .004$ & $.518 \pm .007$ \\
\bottomrule
\end{tabular}
\end{table*}

\section{Ablation Study}
\label{app:ablation}

\begin{figure*}[!tbp]
\centering
\begin{subfigure}[t]{0.32\textwidth}
    \centering
    \includegraphics[width=\textwidth]{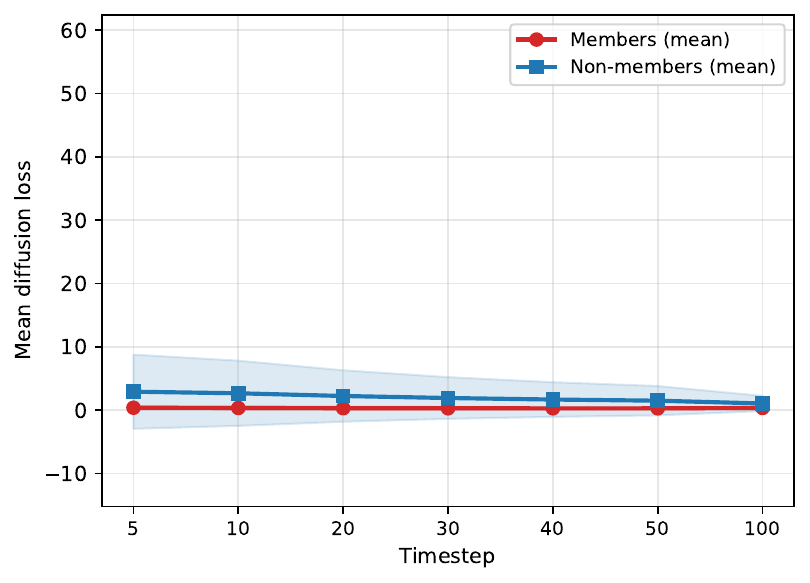}
    \caption{Single-table features}
    \label{fig:loss-single}
\end{subfigure}
\hfill
\begin{subfigure}[t]{0.32\textwidth}
    \centering
    \includegraphics[width=\textwidth]{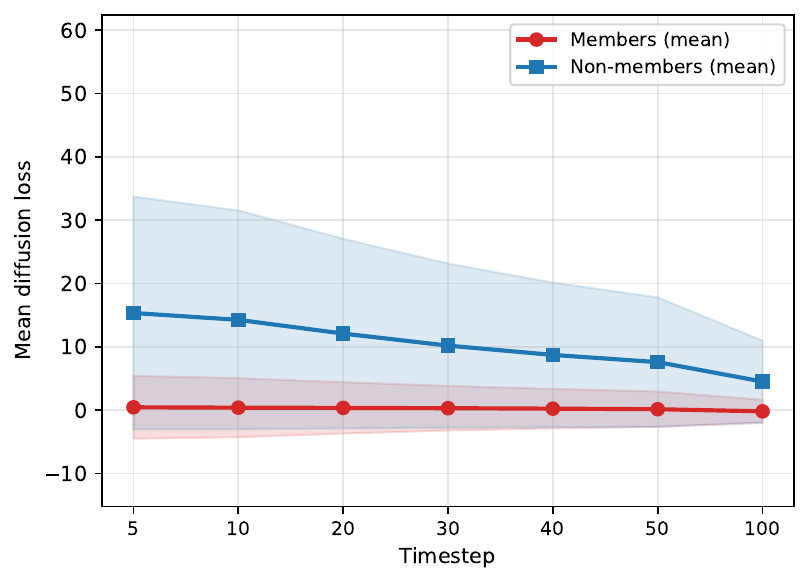}
    \caption{FERMI}
    \label{fig:loss-mapped}
\end{subfigure}
\hfill
\begin{subfigure}[t]{0.32\textwidth}
    \centering
    \includegraphics[width=\textwidth]{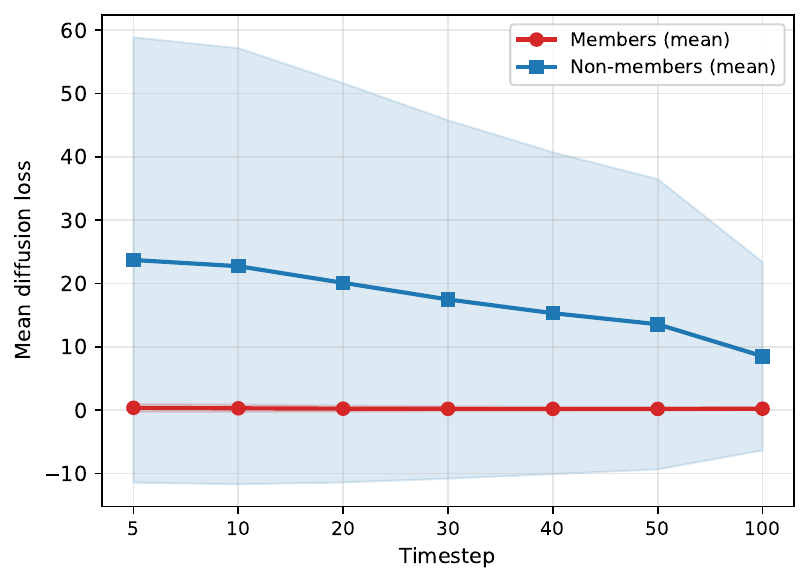}
    \caption{Multi-table features}
    \label{fig:loss-multi}
\end{subfigure}
\caption{Mean diffusion loss per timestep for members and non-members across the three feature settings, for TabDDPM on the Berka dataset. Dots denote mean loss values, and shaded bands indicate $\pm 1$ standard deviation across records.}
\label{fig:features}
\end{figure*}

\paragraph{Membership signal separation across feature regimes.} Figure~\ref{fig:features} illustrates how the discriminative gap between members and non-members evolves across diffusion timesteps under each of the three feature regimes, for TabDDPM trained on the Berka dataset. In the single-table setting (Figure~\ref{fig:features}\subref{fig:loss-single}), training members exhibit consistently near-zero denoising loss across all probed timesteps, reflecting the model's tendency to reconstruct records it has memorized with high fidelity. Non-members display a modestly elevated mean loss, but the separation between the two distributions is narrow. This limited separation explains the moderate single-table attack performance: while a signal exists, it is fragile and difficult to exploit reliably.

The multi-table setting (Figure~\ref{fig:features}\subref{fig:loss-multi}) reveals a fundamentally different picture. By incorporating parent-table context into the feature extraction, the non-member loss curve is substantially amplified, particularly at early timesteps, where the model's sensitivity to relational conditioning is highest. Members continue to exhibit near-zero loss, but the gap between the two distributions widens considerably, and the non-member confidence band shifts upward as a whole. This amplification confirms that relational augmentation exposes a richer membership signal: the model has internalized not only individual record statistics but their relational context, and this joint memorization leaves a stronger fingerprint in the denoising loss profile.

Most critically for our framework, the mapped setting (Figure~\ref{fig:features}\subref{fig:loss-mapped}) occupies a meaningful intermediate position. The feature mapper successfully transfers a portion of this relational amplification into the single-table feature space: non-member loss in the mapped setting is substantially elevated compared to the single-table baseline, particularly at early timesteps, while member loss remains suppressed near zero. Although the mapped curves do not fully replicate the separation magnitude of the true multi-table setting, the mapper demonstrably shifts the feature distribution toward the relational regime, yielding a meaningfully wider and more exploitable member and non-member gap than single-table features alone provide.

Across all three settings, the most discriminative signal concentrates at smaller timesteps, where early-stage denoising errors are most sensitive to whether the record was seen during training. This pattern is consistent with prior findings on loss-fingerprint attacks and suggests that low-noise probing is a particularly effective strategy for membership inference against discrete-time tabular diffusion models~\cite{wu2025winning}.

\paragraph{Does the mapper contribute signal or simply add capacity?}
A natural question arising from the feature mapping framework is whether the observed performance gains stem from the mapper genuinely learning to transfer relational membership signal into the single-table feature space, or whether they are an artifact of additional model capacity and training, that is, whether any neural transformation applied on top of the base features would yield similar improvements. To isolate this, we conduct a controlled ablation in which the mapper is trained not to project single-table features toward their merged-table counterparts, but instead to reconstruct the original single-table features themselves. Under this configuration, the mapper learns an identity-like transformation with minor modifications, and the downstream classifier is trained on its output in exactly the same manner as in the full framework. All architectural and training decisions are held constant; the only meaningful change is the removal of the relational supervision target.

On the California dataset with TabDDPM, this identity-mapping variant achieves an AUC of $0.893 \pm 0.021$, TPR@0.1 of $0.656 \pm 0.032$, and TPR@0.01 of $0.380 \pm 0.142$, statistically indistinguishable from the direct single-table baseline (Table~\ref{tab:single}). Since the architecture and training budget are identical to the full framework, the gains observed under merged-table supervision cannot be attributed to added capacity or extra gradient steps; they arise specifically from the relational supervision signal. When that signal is withheld, the pipeline collapses to the single-table baseline. This validates a core design assumption of the framework: the mapper's contribution is to distill the richer membership information encoded in the merged-table feature space into the single-table regime, rather than to transform features in an arbitrary or capacity-driven way.

\section{Notation}
\label{app:notation}

Table~\ref{tab:notation} summarizes the notation used throughout the paper,
grouped by the section in which each symbol is introduced.

\begin{table}[H]
\caption{Summary of notation used in the paper.}
\label{tab:notation}
\centering
\small
\setlength{\tabcolsep}{4pt}
\begin{tabular}{@{}ll@{\hspace{1.5em}}ll@{}}
\toprule
\textbf{Symbol} & \textbf{Description} & \textbf{Symbol} & \textbf{Description} \\
\midrule
\multicolumn{2}{@{}l}{\textit{Multi-relational schema}} & \multicolumn{2}{@{}l}{\textit{Membership inference}} \\
$\mathcal{D}$ & Multi-relational tabular database & $\mathcal{M}$ & Trained target generative model \\
$M$ & Number of tables in $\mathcal{D}$ & $\mathcal{D}_{\mathrm{train}}(\mathcal{M})$ & Training set of $\mathcal{M}$ \\
$T_j$ & The $j$-th table in $\mathcal{D}$ & $\mathcal{D}_{\mathrm{aux}}$ & Auxiliary dataset \\
$k_j$ & Primary key of table $T_j$ & $\mathcal{D}_{\mathrm{aux}}^{(s)}$ & Auxiliary partition for shadow model $s$ \\
$T_i \to T_j$ & $T_i$ has a foreign key referencing $T_j$ & $x$ & Candidate record under attack \\
$\mathrm{pa}(j)$ & Set of parent tables of $T_j$ & $m(x, \mathcal{M})$ & True membership label, $\in \{0,1\}$ \\
$T_p$ & A parent table, $T_p \in \mathrm{pa}(j)$ & $r(x)$ & Adversary's membership score, $\in [0,1]$ \\
$x_p$ & Record from parent $T_p$ linked to $x$ & $\theta_j$ & Parameters of $p_{\theta_j}$ \\
$x \bowtie x_p$ & Join of $x$ with parent record $x_p$ & $\mathcal{M}^{(s)}$ & The $s$-th shadow model \\
$\tilde{x}$ & Relationally augmented record & $S$ & Number of shadow models \\
$\mathcal{R}$ & Set of related tables forming $\tilde{x}$ & $f_{\mathrm{att}}$ & Attack classifier \\
\cmidrule(r){1-2}
\multicolumn{2}{@{}l}{\textit{Diffusion model}} & PCSI & Parent-Context Side Information \\
$x_0^{(j)}$ & A row sampled from $T_j$ & FRSI & Full-Relational Side Information \\
\cmidrule(r){3-4}
$x_t^{(j)}$ & Noised version of $x_0^{(j)}$ at step $t$ & \multicolumn{2}{@{}l}{\textit{FERMI framework}} \\
$T$ & Total number of diffusion timesteps & $\Phi_{\mathrm{single}}(x)$ & Features from the single-table stream \\
$\beta_t$ & Variance schedule at timestep $t$ & $\Phi_{\mathrm{multi}}(\tilde{x})$ & Features from the relational stream \\
$\alpha_t,\bar{\alpha}_t$ & $\alpha_t = 1-\beta_t$;\ $\bar{\alpha}_t = \prod_{s\le t}\alpha_s$ & $f_{\mathrm{map}}$ & Single-to-multi feature mapper \\
$\epsilon$ & Gaussian noise, $\epsilon \sim \mathcal{N}(0, I)$ & $\hat{x}_{\mathrm{multi}}$ & Mapped features \\
$\epsilon_{\theta_j}$ & Noise-prediction network for $T_j$ & $C_{\mathrm{map}}, C_{\mathrm{multi}}$ & Covariance matrices \\
$p_{\theta_j}$ & Diffusion model trained on $T_j$ & $\mathcal{L}_{\mathrm{map}}$ & Mapper loss (MSE + CORAL) \\
$\mathcal{L}_{\mathrm{diff}}$ & Denoising training objective & $\mathcal{L}_{\mathrm{cls}}$ & Classifier (BCE) loss \\
\cmidrule(r){1-2}
\multicolumn{2}{@{}l}{\textit{Loss-fingerprint features}} & $\mathcal{L}_{\mathrm{total}}$ & Joint fine-tuning loss \\
$\mathcal{T}$ & Set of probed timesteps, $|\mathcal{T}| = n_t$ & $\lambda_{\mathrm{coral}}$ & CORAL alignment weight \\
$\mathcal{E}$ & Set of noise samples, $|\mathcal{E}| = n_\epsilon$ & $\lambda_{\mathrm{map}}, \lambda_{\mathrm{cls}}$ & Joint-loss weights \\
$\ell_{\theta_j}(x_0, t, \epsilon)$ & Per-sample denoising loss & & \\
$\Phi(\cdot)$ & Generic feature extractor & & \\
$d$ & Feature dimension, $d = n_t \cdot n_\epsilon$ & & \\
\bottomrule
\end{tabular}
\end{table}

%%%%%%%%%%%%%%%%%%%%%%%%%%%%%%%%%%%%%%%%%%%%%%%%%%%%%%%%%%%%

\end{document}